\documentclass[sigconf]{acmart}
\renewcommand\footnotetextcopyrightpermission[1]{} 
\AtBeginDocument{%
  \providecommand\BibTeX{{%
    \normalfont B\kern-0.5em{\scshape i\kern-0.25em b}\kern-0.8em\TeX}}}

\copyrightyear{2020}
\acmYear{2020}
\setcopyright{acmlicensed}\acmConference[MM '20]{Proceedings of the 28th ACM International Conference on Multimedia}{October 12--16, 2020}{Seattle, WA, USA}
\acmBooktitle{Proceedings of the 28th ACM International Conference on Multimedia (MM '20), October 12--16, 2020, Seattle, WA, USA}
\acmPrice{15.00}
\acmDOI{10.1145/3394171.3413755}
\acmISBN{978-1-4503-7988-5/20/10}

\acmConference[]{ACM Multimedia}{October 12--16, 2020}{Seattle, WA, USA}
\acmBooktitle{ACM Multimedia
  October 12--16, 2020, Seattle, WA, USA}

\begin{document}
\fancyfoot{}
\fancyhead{}
\title{Language Models as Emotional Classifiers \\ for Textual Conversation}

\author{Connor T. Heaton}
\affiliation{%
  \institution{Pennsylvania State University}
  \city{State College}
  \state{PA}
  \postcode{16802}}
\email{czh5372@psu.edu}

\author{David M. Schwartz}
\affiliation{%
  \institution{Pennsylvania State University}
  \city{State College}
  \state{PA}
  \postcode{16802}}
\email{dms7225@psu.edu}


\begin{abstract}
Emotions play a critical role in our everyday lives by altering how we perceive, process and respond to our environment. Affective computing aims to instill in computers the ability to detect and act on the emotions of human actors. A core aspect of any affective computing system is the classification of a user's emotion. In this study we present a novel methodology for classifying emotion in a conversation. At the backbone of our proposed methodology is a pre-trained Language Model (LM), which is supplemented by a Graph Convolutional Network (GCN) that propagates information over the predicate-argument structure identified in an utterance. We apply our proposed methodology on the IEMOCAP and Friends data sets, achieving state-of-the-art performance on the former and a higher accuracy on certain emotional labels on the latter. Furthermore, we examine the role context plays in our methodology by altering how much of the preceding conversation the model has access to when making a classification.
\end{abstract}





\maketitle

\section{Introduction}

Emotions play a critical role in our everyday lives. They can alter how we perceive, process and respond to our environment. For instance, psychological literature \cite{mikhailova1996abnormal} reveals that people with depression interpret stimuli differently than those without depression. Furthermore, emotions influence how we express ourselves in conversation, revealing more information  than just what was said. Additionally, in conversations, emotional responses can display empathy, communicating understanding and making two (or more) people feel closer together.

Affective computing aims to give computers the ability to detect and act on human emotions. Understanding a user's emotional state provides many new opportunities for computer systems. Detecting frustration could allow a computer to identify when a user is having trouble performing a task and can suggest help, for example. Chatbots used in customer support can engage in more realistic conversations if they are able to understand the emotional content in received messages and incorporate that into a more accurate and meaningful response. Furthermore, affective computing systems can act as a safe guard for individuals with depression; a system that can understand the emotional content of items on the web and the current emotional state of the user can automatically determine when and what it should filter to prevent the user from feeling distressed. 

The first step in all of the systems mentioned above is detecting emotion. This study focuses on the construction of an emotional classifier for textual conversations. Our method leverages pre-trained language models (BERT and XLNet in our experiments) in conjunction with a Graph Convolutional Network (GCN) which is used to process the predicate-argument structure of an utterance, identified through semantic role labeling (SRL). We achieved substantial improvements to the state-of-the-art when applying our method on the IEMOCAP data set. When applying it on the Friends data set, our method does not beat the state-of-the-art, but does a better job of identifying emotions that appear infrequently. We also analyze the importance of context in conversation by altering the amount of preceding utterances the LM's have access to while making their classification.

\section{Related Literature}

As our work deals with conversation transcribed as text, we describe literature pertaining to language modelling first. Then, related work in emotional classification is mentioned. Finally, the data sets being used in the study are discussed.

\subsection{Language Modeling}

Language modeling is a natural language processing (NLP) task in which a model is asked to learn underlying distribution and relation among word tokens in a specified vocabulary \cite{jozefowicz2016exploring}. It is common for a language model (LM) to be \textit{pre-trained} on a large-scale, general purpose corpus before being \textit{fine-tuned} for a specific NLP task. Leveraging a LM in such a fashion has shown to be effective for performing a variety of natural language understanding (NLU) and natural language inference (NLI) tasks including speech recognition \cite{mikolov2010recurrent}, machine translation \cite{schwenk2012large}, and text summarization \cite{filippova2015sentence}.

Until recently most LM architectures were based on recurrent neural networks \cite{jozefowicz2016exploring}. However, in late 2018, Devlin et al released a model they dubbed BERT (\textbf{B}idirectional \textbf{E}ncoder \textbf{R}epresentation from \textbf{T}ransformers) \cite{devlin2018bert}. As the name suggests, BERT is a language model based on the transformer architecture which consists almost entirely of attention modules \cite{vaswani2017attention}. BERT was pre-trained on the BooksCorpus \cite{zhu2015aligning} and English Wikipedia data sets using two pre-training tasks: 1) Masked Language Modeling (MLM) and Next Sentence Prediction (NSP). 

For the MLM pre-training task 15\% of tokens in the input sequence were randomly masked, and BERT was asked to predict the missing tokens. For the NSP pre-training task, two sentences were concatenated together and then processed by BERT as a single input sequence. The \textit{[CLS]} token, a special token prepended to all input sequences, was then extracted from the output and used to make a classification as to whether or not the two sentences appeared next to one another in the source document. BERT achieved state-of-the-art performance on a variety of NLU tasks including the popular GLUE, MultiNLI, and SQuAD benchmarks \cite{devlin2018bert}. 

Shortly after BERT was released, Yang et al proposed XLNet, a generalized autoregressive pre-training method for LM's \cite{yang2019xlnet}. XLNet was similar to BERT in that it was pre-trained on a large, general purpose corpus and based on the transformer architecture, bet differed from BERT on two key respects. First, XLNet was trained to make predictions over all permutations of the input sequence whereas BERT made predictions on the raw input sequence. Second, XLNet leveraged an autoregressive architecture whereas BERT employed an auto-encoder architecture. These modifications break the independence of the tokens in the input sequence assumed by BERT, theoretically allowing XLNet to learn more contextual knowledge. However, the autoregressive nature of XLNet means that generated tokens are only conditioned on tokens up to the current position in the input (to the left) whereas BERT is able to access context from both the left and right of the current position because of it's auto-encoder architecture.

\subsection{Semantic Role Labeling}
Semantic Role Labeling (SRL) is a fundamental NLP task in which a model is asked to identify the predicate-argument structure of a sentence, shedding light on ``who'' did ``what'' to ``whom'' in the input text. To this end, Shi et al proposed a BERT-based model for SRL in 2019 \cite{shi2019simple}. Their model was able to achieve state-of-the-art performance in a variety of SRL benchmarks without the use of external features, such as part-of-speech tags and dependency trees. The choice to not use any external features was a significant departure from previous work on SRL and demonstrated the extent to which BERT is able to model human language. 

\subsection{Graph Neural Networks}
Graph neural networks (GNN's) are a specialized group of deep learning architectures which are able to work with data represented in non-Euclidean domains, such as in a graph \cite{wu2020comprehensive}. While many specialized variants of the GNN have been proposed, one of the most popular GNN architectures is known as a Graph Convolutional Network (GCN) \cite{kipf2016semi}. A GCN is appealing for applications in which data is represented by a graph and embeddings for each node in the graph is desired. GCN's, and many GNN's in general, can be seen as a specialized message passing network in which the new embedding of a node is informed by it's previous embedding, it's neighbors, and it's neighbors' embeddings. 

\subsection{Emotion Classification}

Emotional classification has been done over many modalities. Poria et al. \cite{poria2018multimodal} and Tripathi et al. \cite{tripathi2018multi} performed modality analyses for emotional classification to understand which modes of communication contain the most emotional signal. Their work involved creating many classifiers for different types of data such as video, audio, text, and motion capture. Each model was first trained on data from a single medium as well as different pairings of data from different modalities to compare the performance of unimodal and multimodal models. The two studies found that for unimodal models, performance from best to worst was as follows: text, audio, video, and motion capture. It is worth noting that textual data performed substantially better than the rest (with respect to models made by each researcher). As modalities were combined, model performance continued to increase, as did the computational complexity of the resulting system. A more detailed review of uni- and multimodal emotional models can be found in \cite{poria2017review}.


BERT has been used in both text and auditory systems to classify emotion. The EmotionX 2019 challenge \cite{shmueli2019socialnlp} demonstrates its application in a text-only domain. The challenge was to create an emotional classifier. Eleven teams partook in the competition and seven submitted reports documenting their models. Five of those seven teams used BERT to generate contextual embeddings in their emotion classification system and outperformed the two teams who did not incorporate BERT in their system.

BERT has also been used to classify the emotion of audio in IEmoNet \cite{heusser2019bimodal}. IEmoNet was designed to be a modular system for classifying emotion, where each module can be trained (mostly) independent of the others. The system accepts an audio signal as input and extracts auditory emotional features from it. Concurrently, the audio is transcribed and fed into textual information system. The output from the text system is combined with the audio features extracted earlier and given to a classifier that determines the emotion of the original audio clip. BERT was incorporated into IEmoNet as it had the best performance when classifying based on text only. Furthermore, using BERT as a purely text-based emotion classifier achieved an accuracy only three percentage points lower than the complete IEmoNet model. 

Our research differs from these prior uses of pre-trained LM's towards emotion classification in two key regards. First, we analyze how changing the amount of context given to the model impacts classification accuracy. Second, we supplement the LM with a GCN which generates an additional representation of an utterance, informed by the predicate-argument structure of the utterance identified through SRL. 

\section{Data sets}

During this study, two data sets were used: interactive emotional dyadic motion capture (IEMOCAP) \cite{busso2008iemocap} and Friends \cite{chen2018emotionlines}. The data sets are described in more detail in the sections below.

\subsection{IEMOCAP}

IEMOCAP \cite{busso2008iemocap} captures multimodal emotional data. Ten actors were recruited to perform one-on-one scripted and improvised scenarios designed to show a specific emotion. The dataset includes video and audio recordings, motion capture data of the hands and face, as well as the original scripts and transcribed audio. While scenarios were aimed to elicit specific emotions, the emotional labels in the data set were assigned by six evaluators. 

Evaluators categorized each utterance in the data set into one of ten emotions: \textit{neutral}, \textit{happiness}, \textit{sadness}, \textit{anger}, \textit{surprise}, \textit{fear}, \textit{disgust}, \textit{frustration}, \textit{excited}, and \textit{other}. \textit{Happiness}, \textit{sadness}, \textit{anger}, \textit{disgust}, \textit{fear}, and \textit{surprise} were used as labels because they are considered basic emotions according to \cite{ekman1971constants}. A \textit{neutral} category was added because it was of interest to the creators. Lastly, \textit{frustration} and \textit{excited} labels were added because the creators believed they were important categories to accurately represent the data. 

\begin{figure*}[ht!]
    \centering
    \includegraphics[scale=0.5]{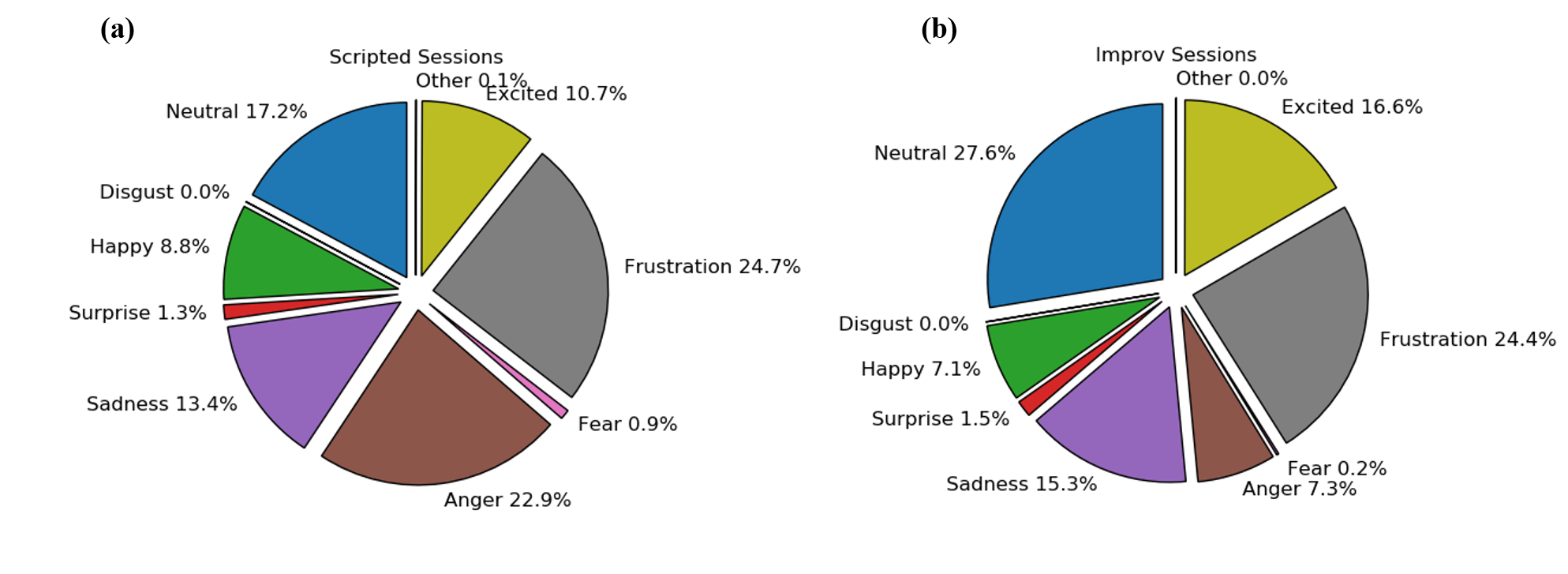}
    \caption{Distribution of labels in datapoints that have a majority agreement on label in IEMOCAP. The pie chart's (a) and (b) represent the distribution of labels for scripted  and improv scenes respectively.}
    \label{fig:IEMOCAP}
\end{figure*}

Each utterance was seen by three different evaluators. The assessments of the displayed emotion from every judge are logged in the data. It is common practice for this data set to filter out data points for which the annotators were unable to reach a consensus. After filtering, about 74\% of the data available for analysis. Figure \ref{fig:IEMOCAP} shows the distribution of labels in scripted and improvised scenes for utterances in the filtered data set. 

Although the IEMOCAP data set contains a total of nine possible emotional labels, researchers using IEMOCAP typically focus on four emotions: \textit{happiness}, \textit{anger}, \textit{sadness}, and \textit{neutral}. 


\subsection{Friends}

EmotionLines \cite{chen2018emotionlines} is a data set containing two smaller sets: EmotionPush and Friends. EmotionPush contains text conversations where each message is labeled with the emotion it projects. The Friends dataset contains scripts from the TV show Friends where each line is labeled with an emotional category. The EmotionX 2019 \cite{shmueli2019socialnlp} challenge had researchers develop emotional classifiers with EmotionLines. After the competition was over, the Friends portion of the data set was posted for others to use. The complete EmotionPush data set can only be obntained via request.

\begin{figure}[ht!]
    \centering
    \includegraphics[scale=0.5]{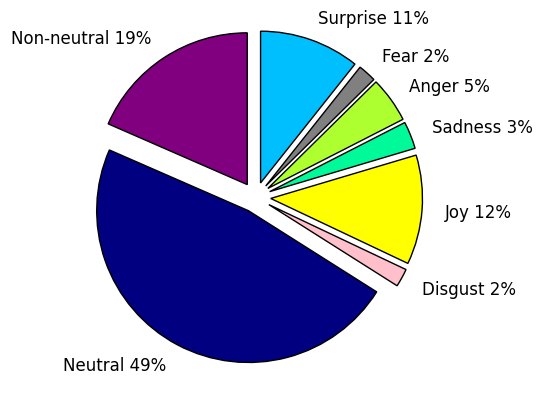}
    \caption{Distribution of labels in the Friends data set.}
    \label{fig:Friends}
\end{figure}

The Friends data set has eight emotional categories: \textit{non-neutral}, \textit{neutral}, \textit{joy}, \textit{sadness}, \textit{anger}, \textit{disgust}, \textit{fear}, and \textit{surprise}. The data was labeled using Amazon Mechanical Turk. Each utterance in the data set was seen by five Turkers. The final emotional label assigned to the utterance was the emotion that received a majority vote by the annotators. The non-neutral category contains all utterances that had no majority vote. Figure \ref{fig:Friends} shows the distribution of utterances and labels in the Friends data set. The figure demonstrates that the data is rather skewed, with the majority being \textit{neutral} utterances. 

The show Friends revolves around six main characters. Thus, unlike in IEMOCAP, the conversations involve more than two people. This makes capturing context in a conversation more difficult. However, it also tests the robustness of the model to generalize to situations involving more than two speakers.

The EmotionX competition asked researchers to create models that categorize an utterance into four categories: \textit{joy}, \textit{sadness}, \textit{anger}, and \textit{neutral}. Therefore, it mimics the outputs typically used by IEMOCAP. However, the EmotionLines paper (\cite{chen2018emotionlines}), that we later compare our method to, performed an analysis on all output categories except non-neutral. As a result, we do not compare to the EmotionX results as we used more output categories. 

\section{Our Methodology}
\begin{figure*}[th!]
    \centering
    \includegraphics[scale=0.50]{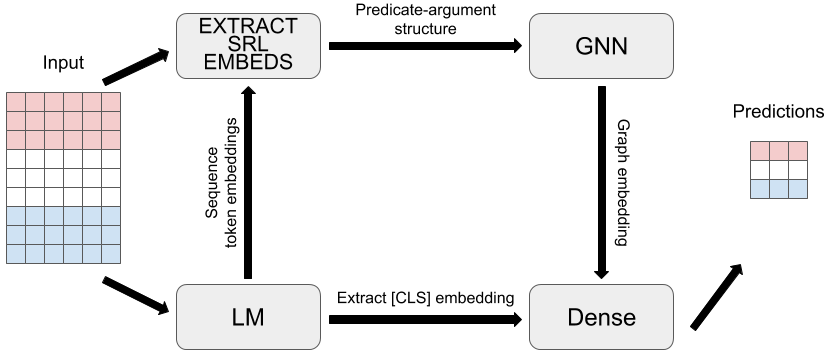}
    \caption{Visual depiction of our proposed methodology. Before being processed by the LM, SRL is performed on the utterance being classified. Once the input sequence has been processed by the LM, embeddings corresponding to elements identified via SRL are extracted and a graph describing the predicate-argument structure is processed by the GNN. Once processed, \textit{multiplicative graph attention} with respect to the processed [CLS] token is used to create an embedding representative of the graph. Finally, a dense layer takes the concatenation of the processed [CLS] token and graph embedding as input to make the final one-versus-all classification.}
    \label{model_architecture}
\end{figure*}

In this section, we describe our methodology in detail, highlighting the importance of each module. The backbone of our proposed methodology can be any pre-trained transformer-based LM, such as BERT \cite{devlin2018bert} or XLNet \cite{yang2019xlnet} described above. Given the relatively small size of most data sets used for identifying emotion in conversation, we believe the use of a LM pre-trained on a general purpose corpus is crucial for achieving a high level of performance. We treat the task of classifying the emotion of an utterance in conversation as a one-versus-all binary classification task for each possible emotion label, and formulate it such that it closely resembles NSP, a task commonly used as a pre-training task for LM's. A visual depiction of our proposed methodology is presented in figure \ref{model_architecture} and described in depth in the following sections.

\subsection{Identifying Predicate-Argument Structure}
The first step in our proposed methodology is the identification of the predicate-argument structure within an individual utterance. Here, we employ the BERT-based SRL model proposed by Shi et al and introduced above \cite{shi2019simple}. We extract the predicates and verbs identified by the BERT-based model in each utterance, disregarding other identified entities. Next, a graph is constructed to represent the predicate-argument structure identified in each utterance. For each predicate-argument set in an utterance, a node is created for both the predicate and argument and an edge is added to connect the two nodes. To increase connectivity of the graph we add an edge between node \textit{a} and node \textit{b} if the set of tokens represented by node \textit{a} are a subset of those represented by node \textit{b}. We note that this process is only performed on the utterance being classified, not any of the utterances provided as context.

\subsection{Constructing the Input Representation}

To assess LM's ability to identify the emotion of an individual utterance in a conversation, as opposed to a single, isolated utterance, we explore the LM's performance when provided with a varying amount of preceding utterances as context. We frame the problem of emotion classification as a one-versus-all binary classification task for each emotion, so an auxiliary sentence must be constructed for all possible emotion labels. The utterance being classified and the preceding \textit{N} context utterances are concatenated together, separated by the \text{[SEP]} token, and used as \textit{text A} for the binary classification. The auxiliary sentence for each possible emotion label takes the form of ``That statement expressed [EMOTION]'' and is used as \textit{text B} for the binary classification. The input sequence to be passed to the LM is formed by concatenating \textit{text A} and \textit{text B} as follows: $\textit{[CLS] text A [SEP] text B [SEP]}.$

\subsection{Fine-Tuning with Graph-Reasoning}
The fine-tuning of vanilla transformer-based LM's such as BERT and XLNet is a relatively straightforward procedure. We utilize pre-trained models provided by Hugging Face\footnote{https://huggingface.co/transformers/} in our experiments. Although a multi-class classification at heart, the LM will first make \textit{E} binary predictions as to whether or not an utterance should be labeled with each of the \textit{E} emotion labels and takes the emotion corresponding to the highest binary prediction as the overall label. 

After the input sequence has been processed, but before each binary prediction is made, a graph is constructed in accordance with the predicate-argument structure identified above. The initial embedding of the $i^{th}$ node, $h_i^0$ is obtained by averaging the embeddings of the corresponding tokens in the LM's output representation. This value is then projected to the graph embedding dimension, $d_{GCN}$, via weight matrix \textit{W}. This process is described below in equation \ref{eq1}.
\begin{equation}\label{eq1}
    h_i^0=\sigma(W\sum_{w_j \in s_i} \frac{1}{|s_i|}h_{w_j})
\end{equation}
where $s_i = \{w_0,...,w_t\}$ are tokens represented by node $i$, $h_{w_j}$ is the LM's contextual representation of the token $w_j$, $W \in R^{d_{LM}xd_{GCN}}$ projects the embeddings, and $\sigma$ is an activation function.

Here, we use a GCN, described above, to process our predicate-argument graph \cite{kipf2016semi}. As such, information propagates through the graph in two phases: aggregation and combination. During aggregation, an intermediate representation of node $i$'s neighbors in layer $l$, $z_i^l$, is influenced by node $i$'s neighbors, $N_i$, and the embedding of those neighbors, $h_j^l$. The aggregation process is described in detail in equation \ref{eq2}, where $V^l$ is the adjacency matrix of the graph in layer \textit{l}.

\begin{equation}\label{eq2}
    z_i^l=\sum_{j \in N_i} \frac{1}{|N_i|} V^l h_j^l
\end{equation}

The combination phase is then executed to obtain a new embedding of node $i$ in layer $l+1$, $h_i^{l+1}$. The new embedding is informed by node $i$'s previous embedding, $h_i^l$, and the intermediate representation of $i$'s neighbors, $z_i^l$. The combination phase is described in detail in equation \ref{eq3} below where $W^l$ is a weight matrix and $\sigma$ an activation function.

\begin{equation}\label{eq3}
    h_i^{l+1}=\sigma(W^l h_i^l + z_i^l)
\end{equation}

Then, multiplicative attention is used to obtain a unified representation of nodes in the graph \cite{luong2015effective}. The embedding of the [CLS] token is extracted from the output of the LM and the attention scores between the [CLS] token and each node in the graph are computed. The unified graph embedding, $h^g$, is then computed as the weighted average of each node in the graph in accordance with the attention scores. This process is described in equations \ref{eq4} and \ref{eq5} below where $h^c$ is the embedding of the [CLS] token, $W_1$ is a weight matrix, $h_i^L$ is the final embedding of node $i$, and $N_i$ is the neighborhood of node $i$.

\begin{equation}\label{eq4}
    \alpha_i = \frac{h^c\sigma(W_1 h_i^l)}{\sum_{j \in N_i}h^c\sigma(W_1h_j^L)}
\end{equation}

\begin{equation}\label{eq5}
    h^g=\sum_{j \in N_i}\alpha_j^L h_j^L
\end{equation}

The unified graph embedding, $h^g$, is then concatenated with the embedding of the [CLS] token, $h^c$, and passed through a dense layer to make the binary classification. The emotion corresponding to the binary prediction for each emotion is then taken as the predicted label for the utterance.

\section{Experiments}
We apply our methodology on both the IEMOCAP and Friends data sets and present our results below. We explore the performance of both \textit{BERT-base-uncased} and \textit{XLNet-base-cased}, the most commonly used variant of each model, in our experiments. In all experiments, the learning rate was set to $5e^{-6}$, a single GCN layer was used, and an Adam optimizer was used with $\beta_1=0.9$ and $\beta_2=0.999$.

\subsection{IEMOCAP}
For experiments on the IEMOCAP data set, utterances for which less than two annotators agreed on a label were excluded. Furthermore, utterances labeled with an emotion other than \textit{anger}, \textit{happiness}, \textit{neutral}, and \textit{sadness} were disregarded as is common practice. Both BERT and XLNet were trained for 9 epochs. Our results obtained by providing BERT and XLNet with 0, 1, 2, 4, and 8 previous utterances as context are presented below in figure \ref{iemocap_metrics}. 

\begin{figure*}[th!]
    \centering
    \includegraphics[scale=0.70]{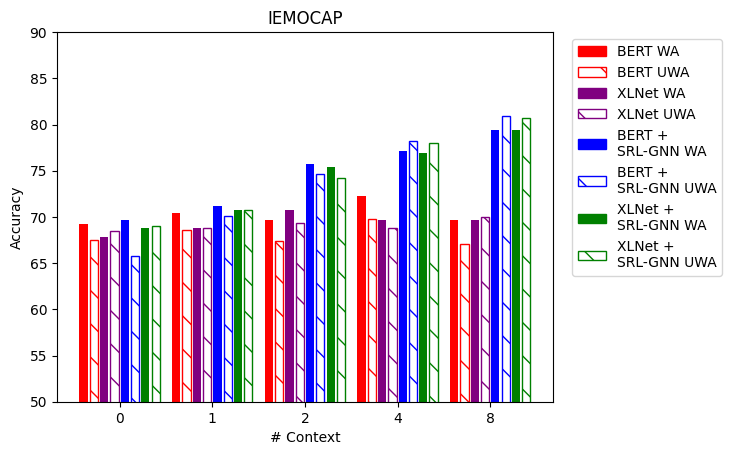}
    \caption{Weighted and un-weighted average accuracy of BERT and XLNet when applied to IEMOCAP data set.}
    \label{iemocap_metrics}
\end{figure*}

Table \ref{Tab:IEMOCAP} compares our model with varying levels of context to prior models mentioned earlier.

\begin{table}[ht]
    \centering
    \begin{tabular}{c | c c }
         Model & WA & UA \\
         \hline
         bc-LSTM \cite{poria2018multimodal} & \multicolumn{2}{c}{73.6\%}  \\   
         LSTM (Text\_Model3) \cite{tripathi2018multi} & \multicolumn{2}{c}{64.78\%} \\
         BERT (portion of IEmoNet) \cite{heusser2019bimodal} & 70.9 \% & 69.1\%  \\
         \hline
         
         BERT+SRL-GNN-1 & 70.10\% & 71.15\% \\
         BERT+SRL-GNN-2 & 74.66\% & 75.77\% \\
         BERT+SRL-GNN-4 & 78.17\% & 77.12\% \\
         \textbf{BERT+SRL-GNN-8} & \textbf{80.90\%} & \textbf{79.42\%} \\
         
         \hline
         
         XLNet+SRL-GNN-1 & 70.76\% & 70.77\% \\
         XLNet+SRL-GNN-2 & 74.16\% & 75.38\% \\
         XLNet+SRL-GNN-4 & 77.98\% & 76.92\% \\
         \textbf{XLNet+SRL-GNN-8} & 80.68\% & \textbf{79.42\%} \\

    \end{tabular}
    \caption{Performance comparison on IEMOCAP. WA represents weighted average accuracy. UA is unweighted average accuracy. Columns with one value reported either the same for WA and UA or did not disclose which metric was reported. The integer at the end of our presented models signifies the number of context utterances provided. }
    \label{Tab:IEMOCAP}
\end{table}

\subsection{Friends}
For experiments on the Friends data set all utterances were considered. Both BERT and XLNet were trained for 11 epochs. Our results obtained by providing BERT and XLNet with 0, 1, 2, 4, and 8 previous utterances as context are presented below in figure \ref{friends_metrics}.

\begin{figure*}[th!]
    \centering
    \includegraphics[scale=0.70]{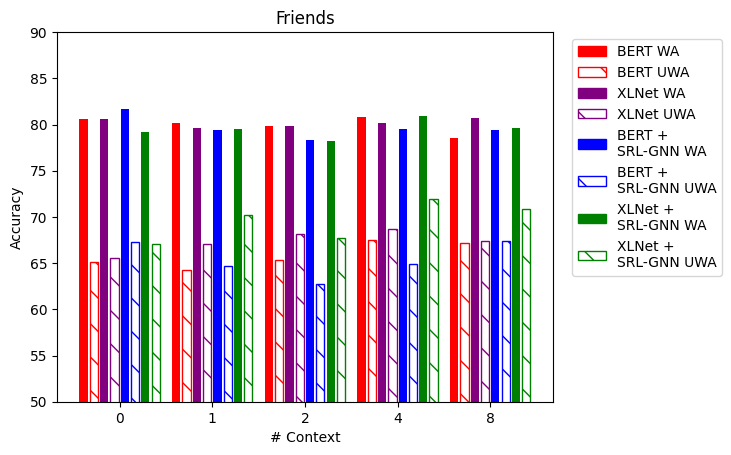}
    \caption{Weighted and un-weighted average accuracy of BERT and XLNet when applied to Friends data set for the four emotions \textit{neutral}, \textit{joy}, \textit{sadness}, and \textit{anger}.}
    \label{friends_metrics}
\end{figure*}

Tables \ref{Tab:Friends_overall} and \ref{Tab:Friends_per_category} compare our models with varying amounts of context over the entire dataset and per category with \cite{chen2018emotionlines}.

\begin{table}[ht]
    \centering
    \begin{tabular}{c | c | c }
         Model & WA & UA  \\
         \hline
         \textbf{CNN-BiLSTM \cite{chen2018emotionlines}} & \textbf{77.4\%} & 39.4\%  \\ 
         \hline
         BERT+SRL-GNN-1 & 70.02\% & 51.63\% \\
         BERT+SRL-GNN-2 & 68.78\% & 49.89\% \\
         BERT+SRL-GNN-4 & 68.78\% & 50.27\% \\
         \textbf{BERT+SRL-GNN-8} & 72.10\% & \textbf{53.71\%} \\

         \hline
         XLNet+SRL-GNN-1 & 69.40\% & 47.68\% \\
         XLNet+SRL-GNN-2 & 71.47\% & 48.23\% \\
         XLNet+SRL-GNN-4 & 71.47\% & 51.36\% \\
         XLNet+SRL-GNN-8 & 72.82\% & 53.41\% \\

    \end{tabular}
    \caption{Performance comparison on Friends dataset. WA is weighted accuracy. UA is unweighted accuracy.}
    \label{Tab:Friends_overall}
\end{table}

\begin{table*}[ht]
    \centering
    \begin{tabular}{ c | c | c | c | c | c | c | c}
        Model & Neu & Joy & Sad & Fea & Ang & Sur & Dis \\
        \hline
        \textbf{CNN-BiLSTM \cite{chen2018emotionlines}} & \textbf{87.0\%} & 60.3\% & 28.7\% & 0.0\% & 32.4\% & 40.9\% & \textbf{26.7\%} \\
        \hline
        \textbf{BERT+SRL-GNN-1} & 82.08\% & 73.98\% & 50.00\% & \textbf{37.93\%} & 41.18\% & 67.55\% & 8.70\% \\
        BERT+SRL-GNN-2 & 81.26\% & 67.48\% & 51.61\% & 27.59\% & 47.06\% & 65.56\% & 8.70\% \\
        BERT+SRL-GNN-4 & 80.86\% & 70.73\% & 54.84\% & 20.69\% & 41.18\% & 66.23\% & 17.39\% \\
        \textbf{BERT+SRL-GNN-8} & 84.32\% & 69.92\% & 48.39\% & 31.03\% & 47.06\% & \textbf{73.51\%} & 21.74\% \\
         
        \hline
        \textbf{XLNet+SRL-GNN-1} & 81.67\% & \textbf{74.80\%} & 46.77\% & 0.00\% & 50.59\% & 66.89\% & 13.04\% \\
        XLNet+SRL-GNN-2 &  85.74\% & 73.98\% & 48.39\% & 0.00\% & 48.24\% & 68.21\% & 13.04\% \\
        XLNet+SRL-GNN-4 & 83.30\% & 73.17\% & 59.68\% & 0.00\% & 54.12\% & 67.55\% & 21.74\% \\
        \textbf{XLNet+SRL-GNN-8} & 85.34\% & 73.98\% & \textbf{67.74\%} & 3.45\% & \textbf{60.00\%} & 61.59\% & 21.74\% \\
    \end{tabular}
    \caption{Performance comparison on Friends dataset per category. The category columns use the first three letters of each label (e.g., neu corresponds to \textit{neutral}).}
    \label{Tab:Friends_per_category}
\end{table*}

\begin{table*}[ht]
    \centering
    \begin{tabular}{ c | c | c | c | c }
        Model & Neu & Joy & Sad & Ang  \\
        \hline
        \textbf{IDEA (BERT-based)\cite{shmueli2019socialnlp}} & 87.30\% & 75.50\% & 59.60\% & \textbf{68.90\%} \\
        KU (BERT-based)\cite{shmueli2019socialnlp} & 86.00\% & 72.00\% & 51.40\% & 65.6\% \\
        HSU (BERT-based)\cite{shmueli2019socialnlp} & 85.40\% & 73.60\% & 55.60\% & 65.00\% \\        
        AlexU (BERT-based)\cite{shmueli2019socialnlp} & 84.50\% & 72.30\% & 58.00\% & 59.70\% \\
        
        \hline
        \textbf{BERT+SRL-GNN-0} & \textbf{93.08\%} & 69.92\% & 48.39\% & 57.65\% \\
        BERT+SRL-GNN-1 & 91.45\% & 65.85\% & 53.23\% & 48.24\% \\
        BERT+SRL-GNN-2 & 91.45\% & 62.60\% & 53.23\% & 43.53\% \\
        BERT+SRL-GNN-4 & 90.84\% & 69.11\% & 46.77\% & 52.94\% \\
        BERT+SRL-GNN-8 & 88.80\% & 69.92\% & 53.23\% & 57.65\% \\
         
        \hline
        \textbf{XLNet+SRL-GNN-0} & 87.37\% & \textbf{80.49\%} & 45.16\% & 55.29\% \\
        XLNet+SRL-GNN-1 & 86.35\% & 77.24\% & 59.68\% & 57.65\% \\
        XLNet+SRL-GNN-2 & 86.56\% & 71.54\% & 59.68\% & 52.94\% \\
        XLNet+SRL-GNN-4 & 87.78\% & 77.24\% & 62.90\% & 60.00\% \\
        \textbf{XLNet+SRL-GNN-8} & 86.76\% & 75.61\% & \textbf{69.35\%} & 51.76\% \\
    \end{tabular}
    \caption{Performance comparison on Friends dataset per category. The category columns use the first three letters of each label (e.g., neu corresponds to \textit{neutral}).}
    \label{Tab:Friends_per_category}
\end{table*}

\section{Discussion}
Upon inspecting the results of our experiments we generally see an increase in performance for both BERT and XLNet when applied to the IEMOCAP and Friends data sets when our methodology is used. However, the realized increase in performance varies by model and data set. When preceding utterances are not provided as context, models trained both with and without our methodology achieve similar levels of performance. We hypothesize that this is because both models are able to identify the relevant predicate-argument structure in an utterance without the use of a GCN. Recent work presented by Clark et al suggests that different attention heads in BERT are able to attend to specific types of tokens, such as direct objects, noun modifiers, passive auxiliary verbs, and prepositions, among others, lending credence to our hypothesis \cite{clark2019does}. While a similar analysis has not been done for XLNet, due to their very similar architecture, we believe it is not unreasonable to assume the attention heads in XLNet perform a similar function. 

\subsection{IEMOCAP}
We see very strong performance gains when our methodology is applied to the IEMOCAP data set. Performance of the vanilla XLNet and BERT models peak when two and four utterances are provided as context, respectively, but degrades when eight utterances are provided as context. When our methodology is used, however, we see performance continue to increase as more utterances are provided as context. We hypothesize that the performance of the vanilla models degrades when more than four utterances are provided as context because the length of the input sequence approaches the maximum input length of both BERT and XLNet. As a result, we believe the models struggle to identify important syntactic and semantic information in each utterance as well as they can when dealing with shorter input sequences. The inclusion of our proposed GCN module appears to alleviate the issue encountered when dealing with longer sequences. 

\subsection{Friends}
The results of applying our methodology to the Friends data set do not tell as definitive a story as the IEMOCAP results. We do not see a clear difference in the performance of either XLNet or BERT when our methodology is employed, nor do we see a trend in performance as the number of utterances provided as context increases. However, in general, our methodology does seem to provide an increase in performance for particular experimental settings. For example, when four utterances are provided as context, BERT sees roughly a 5\% increase in unweighted average accuracy by using our methodology, but XLNet achieves the same level of performance whether our methodology is used or not. Furthermore, we observe improved unweighted accuracy metrics when our model is applied with both LM's compared to when it is not applied. This would suggest that our model is not as sensitive to the distribution of labels in the data set compared to previous methods \cite{shmueli2019socialnlp}.

One possible explanation for the inconsistent performance on the Friends data set is that the models struggled to handle conversations containing multiple speakers. All conversations in the IEMOCAP data set involve \textit{exactly} two participants, while conversations in the Friends data set all contain \textit{at least} two participants. Another likely cause for the inconsistent performance is the imbalanced nature of the data set. For example, utterances with the \textit{neutral} label account for roughly 45\% of the data set while only 2\% of utterances are labeled as \textit{fear}.

\section{Future Work}
While we showed that our methodology provides significant improvements over previous state-of-the-art on the IEMOCAP data set, we believe even higher levels of performance could be reached with a few modifications to our framework. We briefly describe these possible future directions in the sections below. 

\subsection{Fully Exploiting LM's}
In our experiments we exclusively took the output from BERT and XLNet's final layer as the embedding for tokens in our input sequence. However, the output is actually the $13^{th}$ representation of the input sequence produced by the LM's (recall that each LM is formed by stacking $N$ transformer modules, $N=12$ for the \textit{base} models used in our experiments). Additionally, the models compute an attention matrix for the input sequence in each layer. We believe these 12 internal embeddings and attention matrices may contain additional emotional signal which could be used to further improve performance. Clark et al have recently demonstrated how particular layers in BERT, and specific attention heads in different layers, have different functionality, perhaps suggesting this is a fruitful path forward \cite{clark2019does}.

\subsection{Better Leveraging Context Utterances}
Another interesting path forward could include identifying the predicate-argument structure in utterances in the context, as well as in the utterance being classified. Having shown that performance improves when SRL is performed on only the utterance being classified, it is not outlandish to assume that the performing the same process on context utterances could further improve performance.

\subsection{More Accurately Modeling Group Conversation}
As mentioned above, we believe the multi-party nature (>2 participants) of the conversations in the Friends data set is partially to blame for the inconsistent performance we observed. A brief inspection of the data revealed that in some conversations, a subset of participants will go off on a tangent, straying from the rest of the group. Not only does the model need to handle multiple participants, but also multiple topics of conversation. To this end, we believe it would be interesting to explore the effect of introducing \textit{participant-specific} special tokens to our model, one for each participant. We believe this may allow the LM's to better manage multiple participants, and perhaps even multiple topics of conversation. 

\section{Conclusion}
Emotions play a crucial role in our everyday lives. Computers can benefit from having the ability to detect and act on the emotional state of its user. To act on emotions, the user's state must first be known.
Our experiments show pre-trained LM's, such as BERT and XLNet, can be used for the classification of emotion in conversation. We supplement the LM's by extracting the predicate-argument structure of the utterance being classified as a graph, and create a unified graph representation through the use of a GCN and multiplicative graph attention. This graph representation is then used as additional input to the models when making a classification for each emotion. We see substantial performance improvements when our methodology is applied to the IEMOCAP data set, but performance when applied to the Friends data set is much more variable.

\bibliographystyle{acm}
\bibliography{sample-base}

\begin{thebibliography}{10}

\bibitem{busso2008iemocap}
{\sc Busso, C., Bulut, M., Lee, C.-C., Kazemzadeh, A., Mower, E., Kim, S.,
  Chang, J.~N., Lee, S., and Narayanan, S.~S.}
\newblock Iemocap: Interactive emotional dyadic motion capture database.
\newblock {\em Language resources and evaluation 42}, 4 (2008), 335.

\bibitem{chen2018emotionlines}
{\sc Chen, S.-Y., Hsu, C.-C., Kuo, C.-C., Ku, L.-W., et~al.}
\newblock Emotionlines: An emotion corpus of multi-party conversations.
\newblock {\em arXiv preprint arXiv:1802.08379\/} (2018).

\bibitem{clark2019does}
{\sc Clark, K., Khandelwal, U., Levy, O., and Manning, C.~D.}
\newblock What does bert look at? an analysis of bert's attention.
\newblock {\em arXiv preprint arXiv:1906.04341\/} (2019).

\bibitem{devlin2018bert}
{\sc Devlin, J., Chang, M.-W., Lee, K., and Toutanova, K.}
\newblock Bert: Pre-training of deep bidirectional transformers for language
  understanding.
\newblock {\em arXiv preprint arXiv:1810.04805\/} (2018).

\bibitem{ekman1971constants}
{\sc Ekman, P., and Friesen, W.~V.}
\newblock Constants across cultures in the face and emotion.
\newblock {\em Journal of personality and social psychology 17}, 2 (1971), 124.

\bibitem{filippova2015sentence}
{\sc Filippova, K., Alfonseca, E., Colmenares, C.~A., Kaiser, {\L}., and
  Vinyals, O.}
\newblock Sentence compression by deletion with lstms.
\newblock In {\em Proceedings of the 2015 Conference on Empirical Methods in
  Natural Language Processing\/} (2015), pp.~360--368.

\bibitem{heusser2019bimodal}
{\sc Heusser, V., Freymuth, N., Constantin, S., and Waibel, A.}
\newblock Bimodal speech emotion recognition using pre-trained language models.
\newblock {\em arXiv preprint arXiv:1912.02610\/} (2019).

\bibitem{jozefowicz2016exploring}
{\sc Jozefowicz, R., Vinyals, O., Schuster, M., Shazeer, N., and Wu, Y.}
\newblock Exploring the limits of language modeling.
\newblock {\em arXiv preprint arXiv:1602.02410\/} (2016).

\bibitem{kipf2016semi}
{\sc Kipf, T.~N., and Welling, M.}
\newblock Semi-supervised classification with graph convolutional networks.
\newblock {\em arXiv preprint arXiv:1609.02907\/} (2016).

\bibitem{luong2015effective}
{\sc Luong, M.-T., Pham, H., and Manning, C.~D.}
\newblock Effective approaches to attention-based neural machine translation.
\newblock {\em arXiv preprint arXiv:1508.04025\/} (2015).

\bibitem{mikhailova1996abnormal}
{\sc Mikhailova, E.~S., Vladimirova, T.~V., Iznak, A.~F., Tsusulkovskaya,
  E.~J., and Sushko, N.~V.}
\newblock Abnormal recognition of facial expression of emotions in depressed
  patients with major depression disorder and schizotypal personality disorder.
\newblock {\em Biological psychiatry 40}, 8 (1996), 697--705.

\bibitem{mikolov2010recurrent}
{\sc Mikolov, T., Karafi{\'a}t, M., Burget, L., {\v{C}}ernock{\`y}, J., and
  Khudanpur, S.}
\newblock Recurrent neural network based language model.
\newblock In {\em Eleventh annual conference of the international speech
  communication association\/} (2010).

\bibitem{poria2017review}
{\sc Poria, S., Cambria, E., Bajpai, R., and Hussain, A.}
\newblock A review of affective computing: From unimodal analysis to multimodal
  fusion.
\newblock {\em Information Fusion 37\/} (2017), 98--125.

\bibitem{poria2018multimodal}
{\sc Poria, S., Majumder, N., Hazarika, D., Cambria, E., Gelbukh, A., and
  Hussain, A.}
\newblock Multimodal sentiment analysis: Addressing key issues and setting up
  the baselines.
\newblock {\em IEEE Intelligent Systems 33}, 6 (2018), 17--25.

\bibitem{schwenk2012large}
{\sc Schwenk, H., Rousseau, A., and Attik, M.}
\newblock Large, pruned or continuous space language models on a gpu for
  statistical machine translation.
\newblock In {\em Proceedings of the NAACL-HLT 2012 Workshop: Will We Ever
  Really Replace the N-gram Model? On the Future of Language Modeling for
  HLT\/} (2012), Association for Computational Linguistics, pp.~11--19.

\bibitem{shi2019simple}
{\sc Shi, P., and Lin, J.}
\newblock Simple bert models for relation extraction and semantic role
  labeling.
\newblock {\em arXiv preprint arXiv:1904.05255\/} (2019).

\bibitem{shmueli2019socialnlp}
{\sc Shmueli, B., and Ku, L.-W.}
\newblock Socialnlp emotionx 2019 challenge overview: Predicting emotions in
  spoken dialogues and chats.
\newblock {\em arXiv preprint arXiv:1909.07734\/} (2019).

\bibitem{tripathi2018multi}
{\sc Tripathi, S., Tripathi, S., and Beigi, H.}
\newblock Multi-modal emotion recognition on iemocap with neural networks.
\newblock {\em arXiv preprint arXiv:1804.05788\/} (2018).

\bibitem{vaswani2017attention}
{\sc Vaswani, A., Shazeer, N., Parmar, N., Uszkoreit, J., Jones, L., Gomez,
  A.~N., Kaiser, {\L}., and Polosukhin, I.}
\newblock Attention is all you need.
\newblock In {\em Advances in neural information processing systems\/} (2017),
  pp.~5998--6008.

\bibitem{wu2020comprehensive}
{\sc Wu, Z., Pan, S., Chen, F., Long, G., Zhang, C., and Philip, S.~Y.}
\newblock A comprehensive survey on graph neural networks.
\newblock {\em IEEE Transactions on Neural Networks and Learning Systems\/}
  (2020).

\bibitem{yang2019xlnet}
{\sc Yang, Z., Dai, Z., Yang, Y., Carbonell, J., Salakhutdinov, R.~R., and Le,
  Q.~V.}
\newblock Xlnet: Generalized autoregressive pretraining for language
  understanding.
\newblock In {\em Advances in neural information processing systems\/} (2019),
  pp.~5754--5764.

\bibitem{zhu2015aligning}
{\sc Zhu, Y., Kiros, R., Zemel, R., Salakhutdinov, R., Urtasun, R., Torralba,
  A., and Fidler, S.}
\newblock Aligning books and movies: Towards story-like visual explanations by
  watching movies and reading books.
\newblock In {\em Proceedings of the IEEE international conference on computer
  vision\/} (2015), pp.~19--27.

\end{thebibliography}

\end{document}